\documentclass[conference]{IEEEtran}
\IEEEoverridecommandlockouts
\usepackage{cite}
\usepackage{amsmath,amssymb,amsfonts}
\usepackage{algorithmic}
\usepackage{graphicx}
\usepackage{fancyvrb}
\usepackage{caption} 
\usepackage{alltt}
\usepackage{textcomp}
\usepackage{xcolor}
\def\BibTeX{{\rm B\kern-.05em{\sc i\kern-.025em b}\kern-.08em
    T\kern-.1667em\lower.7ex\hbox{E}\kern-.125emX}}
\usepackage{balance}
\usepackage{lastpage}
\emergencystretch=\maxdimen
\begin{document}
\title{A Deep Learning Approach for Ontology Enrichment from Unstructured Text}

\author{\IEEEauthorblockN{Lalit Mohan Sanagavarapu\textsuperscript{\textsection}}
\IEEEauthorblockA{\textit{IIIT Hyderabad}\\
Hyderabad, India \\
lalit.mohan@research.iiit.ac.in}
\and
\IEEEauthorblockN{Vivek Iyer\textsuperscript{\textsection}}
\IEEEauthorblockA{\textit{IIIT Hyderabad}\\
Hyderabad, India \\
vivek.iyer@research.iiit.ac.in}
\and
\IEEEauthorblockN{Y Raghu Reddy}
\IEEEauthorblockA{\textit{IIIT Hyderabad}\\
Hyderabad, India \\
raghu.reddy@iiit.ac.in}
}

\maketitle
\begingroup\renewcommand\thefootnote{\textsection}
\footnotetext{Equal contribution}
\endgroup

\begin{abstract}
Information Security in the cyber world is a major cause for concern, with significant increase in the number of attack surfaces. Existing information on vulnerabilities, attacks, controls, and advisories available on web provides an opportunity to represent knowledge and perform security analytics to mitigate some of the concerns. Representing security knowledge in the form of ontology facilitates anomaly detection, threat intelligence, reasoning and relevance attribution of attacks, and many more. This necessitates dynamic and automated enrichment of information security ontologies. However, existing ontology enrichment algorithms based on natural language processing and ML models have issues with contextual extraction of concepts in words, phrases and sentences. This motivates the need for sequential Deep Learning architectures that traverse through dependency paths in text and extract embedded vulnerabilities, threats, controls, products and other security related concepts and instances from learned path representations. In the proposed approach, Bidirectional LSTMs trained on a large DBpedia dataset and Wikipedia corpus of 2.8 GB along with Universal Sentence Encoder is deployed to enrich ISO 27001 \cite{SecKnowlFenz} based information security ontology. The model is trained and tested on an high performance computing (HPC) environment to handle Wiki text dimensionality. The approach yielded a test accuracy of over 80\% when tested with knocked out concepts from ontology and web page instances to validate the robustness.
\end{abstract}


\section{Introduction}
\begin{figure*} [h]
    \centering
    \includegraphics[width=1.0\linewidth]{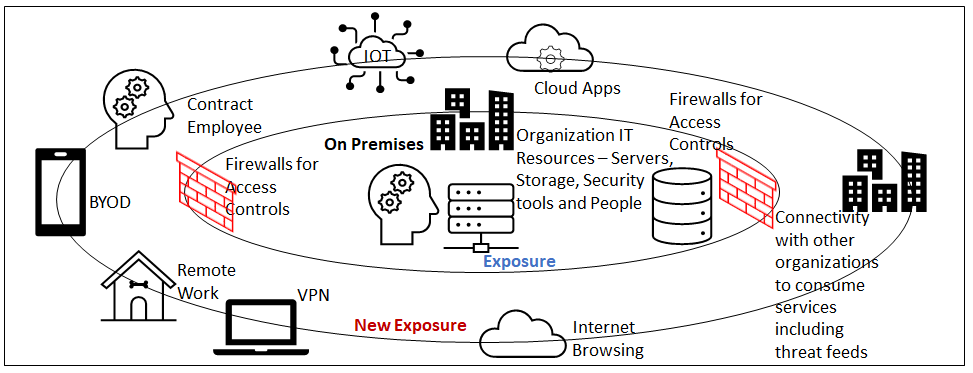}
    \caption{Changing Attack Surface}
    \label{fig:attacksurface}
\end{figure*}

In recent times, there is an exponential increase in the number of content providers and content consumers on the internet due to various reasons like improved digital literacy, affordable devices, better network, etc. Further, the number of internet-connected devices per person is expected to increase even more with adoption of emerging technologies such as Internet of Things and 5G. This change in users and usage is leading to an increase in data breaches\footnote{https://digitalguardian.com/blog/history-data-breaches}. In many cases, realization of an impact happens long after the attack. The Fig. \ref{fig:attacksurface} shows changing attack surface for organizations with remote work force and connected devices.

Typically, organizations invest in security tools and infrastructure that are based on rules, statistical models and machine learning (ML) techniques to identify and mitigate the risks arising from the threats. Firewalls, intrusion detection and prevention systems, authentication and authorization mechanisms to data and servers, encryption layers, anti-virus software, endpoint controls and permissions are some of the tools and processes are used to protect Information Technology (IT) systems. Apart from these controls and processes, IT systems are regularly patched to mitigate the risks. 

In addition, organizations purchase threat intelligence feeds to continuously monitor IT infrastructure for anomaly detection. The subscription fee of threat intelligence feeds from service providers is expensive and to a large extent, it contains threat intelligence that is already available in public forums. Public forums such as blogs, discussion forums, government sites, social media channels including Twitter and others contain unstructured threat intelligence on vulnerabilities, attacks, and controls. Tech-savvy internet users interested in information security access public forums, search, and browse on security products, their configurations, reviews, vulnerabilities and other related content for awareness and to protect IT assets.

In recent years, organization's information security infrastructure use Structured Threat Information eXchange (STIX) / Trusted Automated Exchange of Intelligence Information (TAXII) knowledge representation from OASIS\footnote{https://www.oasis-open.org/} to represent observable objects and their properties in the cyber domain. However, automated processing of unstructured text to generate STIX format is a formidable challenge \cite{syed2016uco}. Interestingly, there are transformations available to convert from XML based STIX format to ontological `OWL' or `RDF' formats, which in part, has influenced OASIS to adopt ontology for representations. 

Evidently, the research to use unstructured security related content to enrich ontologies is gaining ground to mitigate risks related to zero-day attacks, malware characterization, digital forensics and incidence response and management \cite{syed2016uco,iannacone2015developing,al2019automatic}. Security ontologies are used to analyze vulnerabilities and model attacks \cite{SecKnowlFenz,jia2018practical,iannacone2015developing,zheng2018learning}. The concepts, relationships and instances of security ontologies are used to validate level of defence-in-depth to protect IT assets, map security product features to controls which leads to assurance of the security infrastructure. The constraints and properties of ontologies allow root cause analysis of attacks. Additionally, given that security-related data is in structured, semi-structured or unstructured forms, unifying them with ontologies aids in situational awareness and readiness to defend an attack \cite{syed2016uco}.
%


Traditionally, domain experts constructed and maintained ontologies. Given the extent of effort and cost involved, access to domain content and ability to process text with advanced natural language processing (NLP) techniques and ML models on powerful IT infrastructure opens up research opportunities to construct and manage ontologies. The information security ontologies can be constructed or enriched from unstructured text available on public forums, vulnerability databases such as National Vulnerability Database (NVD) \footnote{https://nvd.nist.gov} and other information security processing systems \cite{sanagavarapu2018siren, AlienURL} sources. Also, standards and guidelines from ISO/IEC \cite{ISO27001}, NIST from US, ENISA from European Nation, Cloud Security Alliance (CSA) and others to protect confidentiality, integrity and availability of IT assets, contain embedded concepts. The ISO 27001:2015 \cite{SecKnowlFenz} based security ontologies that encompass most of these guidelines are being extensively explored for protection, auditing and compliance checking. Hence, enrichment of ISO 27001 based ontology provides wider acceptance, easier management and interoperability. %

In this work, we propose to enrich a widely accepted Information Security ontology instead of constructing a new ontology from text. This avoids inclusion of trivial concepts and relations. The success of enrichment also enables wider acceptance and usage by domain experts. However, the available literature on ontology enrichment from text is based on approaches utilizing word similarity and supervised ML models \cite{iannacone2015developing, sayan2019semantic}. These ontology enrichment approaches, albeit useful to extract word-level concepts, are limited with respect to (a) extraction of longer concepts embedded in compound words and phrases (b) factoring context while identifying relevant concepts and (c) extracting and classifying instances \cite{iyer2019survey}. 

In the proposed approach ($OntoEnricher$), we implemented a supervised sequential deep learning model that: a) factors context from grammatical and linguistic information encoded in the dependency paths of a sentence, and then b) utilizes sequential neural networks, such as Bidirectional Long Short Term Memory (LSTM) \cite{650093} to traverse (forward and backward directions) dependency paths and learn relevant path representations that constitute relations. Bidirectional LSTM model has ability to forget unrelated stream of data to identify related concepts that are available in the form of a word, a phrase or a sentence in the text. In addition, we utilized pre-trained transformer-based architecture of Universal Sentence Encoder (USE) \cite{cer2018universal} to handle distributional representations of compound words, phrases, and instances. 

The proposed $OntoEnricher$ is implemented on Information security ontology. As availability of information security datasets is a concern, a semi-automatic approach with a training dataset of 97,425 related terms (hypernyms, hyponyms and instances) is extracted from DBpedia for all concepts of a information security ontology \cite{SecKnowlFenz}. To learn syntactic and semantic dependency structure in sentences, a 2.6 GB training corpus on information security is extracted from Wikipedia of all terms in the ontology and the DBpedia dataset. The curated dataset and corpus are used to train bidirectional LSTM model in the proposed ontology enrichment approach. The trained model is tested to enrich concepts, relations, and instances in information security ontology from unstructured text on the internet. The $OntoEnricher$ is also tested with 10\% of training dataset, knocking out terms from ontology and unstructured text from web pages and achieved an average accuracy of 80\%, which is better than current state-of-the-art approaches. As the text in corpus is multi-dimensional and dependency path gets generated for very matching pair of dataset terms, we used a high performance computing (HPC) cluster for training and testing of model faster \cite{lim2019methods}. The code and documentation of ontology enrichment pipeline are publicly available on GitHub for reuse and extension. The subsequent sections in this chapter includes (a) an elaboration of $OntoEnricher$ approach along with an example; (b) Experiment and Results; (c) Discussion and potential future work.
\section{Related Work}
This section discusses related work on enrichment of ontologies from unstructured text as well as approaches to create and maintain information security ontologies. The work on enrichment of knowledge graphs (KG) from unstructured text is also discussed as it represents knowledge and contains similarities with ontologies.
%

Researchers worked on knowledge acquisition from text to construct ontologies for past couple of decades \cite{buitelaar2005ontology, liu2011natural}. The last decade witnessed significant progress in the field of information extraction from web with projects such as DBpedia, Freebase and others. The work of Mitchell et. al \cite{mitchell2018never} known as `NELL' states that it is a never-ending system to learn from web, their work bootstraps knowledge graphs on a continuous basis. Tools such as ReVerb \cite{fader2011identifying} and OLLIE \cite{schmitz2012open} are based on open information systems to extract a triple from a sentence using syntactic and lexical patterns. Although these approaches extract triples from unstructured text using shallow and fast models, they do not handle ambiguity while entity mapping and do not learn expressive features compared to deep and multi-layer models. 

The ML models based on probabilistic, neural networks and others are also explored for ontology enrichment from text \cite{liu2011natural, petasis2011ontology, pingle2019relext}. In 2017, Wang et al \cite{wang2017knowledge} conducted a survey on knowledge graph completion, entity classification and resolution, and relation extraction. The study classified embedding techniques into translational distance models and semantic matching models. The study also stated that additional information in the form of entity types, textual descriptions, relation paths and logical rules strengthen the research. Deep learning models such as CNN \cite{dettmers2018convolutional}, LSTM \cite{nie2019knowledge, li2019biomedical} and variants are used to construct knowledge graphs from text as they carry memory cells and forget gates to build the context and reduce noise. The work of Vedula et al. \cite{vedula2019bolt} proposed an approach to bootstrap newer ontologies from related domains.

Some of the recent approaches are based on Word2Vec  \cite{wohlgenannt2016using} and its variants such as Phrase2Vec or Doc2Vec that use distributional similarities to identify concepts to enrich an ontology. However, these approaches underperform in the extraction of concepts embedded in words, phrases and sentences due to their inability to adequately characterize context. Compared to Word2Vec and its variants, Universal Sentence Encoder (USE) \cite{cer2018universal} stands promising to identify concepts in long phrases as it encodes text into high dimensional vectors for semantic similarity. Lately, researchers \cite{ganesan2020neural} are exploring USE to produce sentence embeddings and deduce semantic closeness in queries. Although, transformer-based models such as BERT and XLNet \cite{liu2020survey, ezen2020comparison} are of interest to ontology enrichment researchers, training them to a domain is effort intensive. 

The literature to enrich security ontologies from text drew attention with OASIS's STIX/TAXII standardization and open source threat intelligence. Most of the current work on security ontologies from text (construction or enrichment) are based on usage of string, substring, pre-fix and post-fix matching of terms, Word2Vec and other basic ML models \cite{obrst2012developing, syed2016uco, pingle2019relext}. In ontologies as well, the deep learning approaches based on recurrent neural networks are trending because of their ability to build the context over multiple words \cite{jia2018practical, gasmi2019cold}. The research of Houssem et al. \cite{gasmi2019cold} used LSTM for population of security ontologies. However, the details to create corpus, handle phrases and robustness of the approach are not elaborated, only 40 entities are used in the model. The literature revealed that security ontologies based on ISO 27001 \cite{SecKnowlFenz} and MITRE Corporation's cyber security effort \cite{syed2016uco} are most referred.
\section{Ontology Enrichment Approach}
In the proposed approach, whenever a new concept is introduced, current memory state of LSTMs are updated to replace old concept, or add new concept by multiplying with forget gates as needed. The concept in current memory are mapped to instances and relations between concepts in current state are constructed. Concepts extracted are used to update the ontology automatically or after manual validation by a domain expert.

\begin{figure*} [h]
    \centering
    \includegraphics[width=1.0\linewidth]{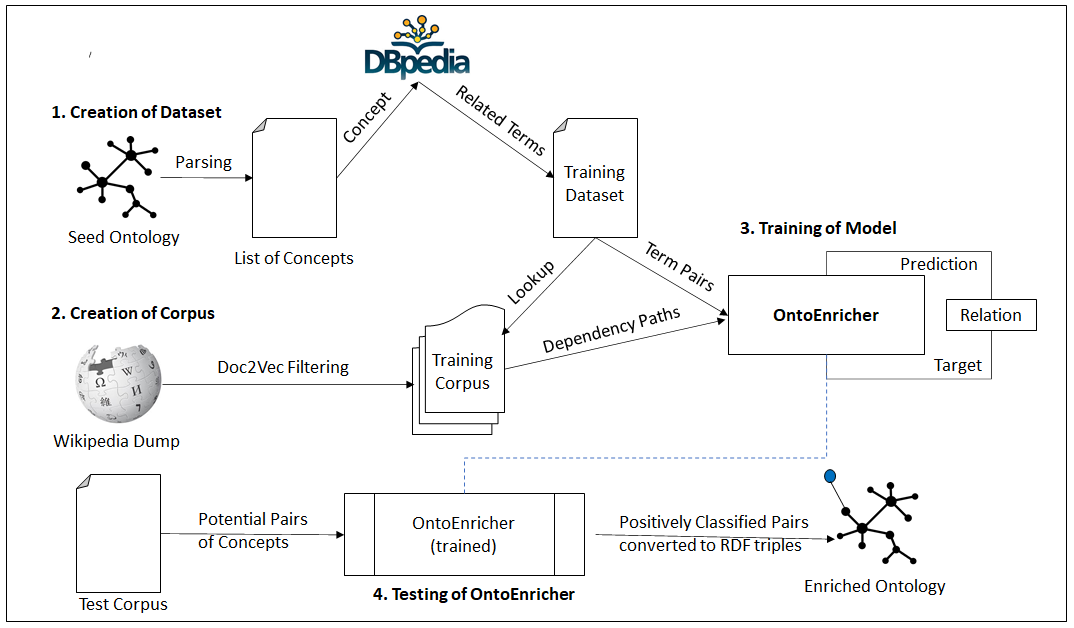}
    \caption{Approach used for Ontology Enrichment}
    \label{fig:EnrichApproach}
\end{figure*}

The $OntoEnricher$ enriches a seed ontology with concepts, relations and instances extracted from unstructured text. As shown in Fig. \ref{fig:EnrichApproach}, the ontology enrichment approach consists of four stages: (i) $Dataset Creation:$ creates training dataset by extracting and curating related terms from DBpedia for all concepts in the ontology (ii) $Corpus Creation:$ creates domain-specific training corpus by parsing Wikipedia dump using various filtering measures (iii) $Training:$ trains $OntoEnricher$ for relation classification of term pairs using training dataset and corpus, and (iv) $Testing:$ tests the approach by enriching the ontology from domain-specific web pages. 
\subsubsection{Stage 1: Creation of Dataset}
\begin{figure} [h]
    \centering
    \includegraphics[width=1.0\linewidth,keepaspectratio]{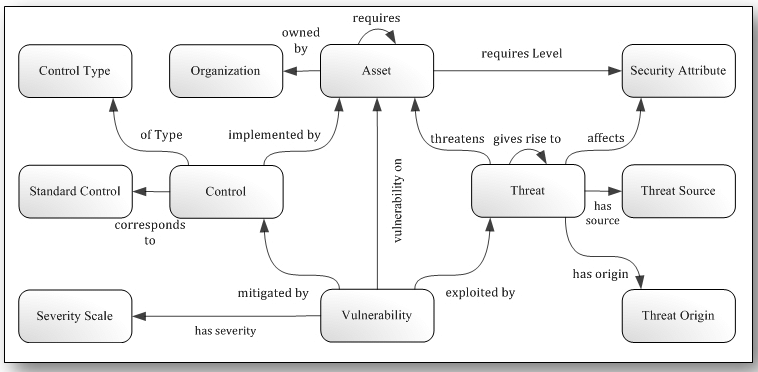}
    \caption{Information Security Upper Ontology \cite{SecKnowlFenz}}
    \label{fig:SecUppOnto}
\end{figure}

The information security seed ontology is based on ISO 27001 \cite{SecKnowlFenz}. The standard ISO 27001:2015 \cite{ISO27001} contains 114 controls across 14 groups. These groups are `Human Resources', `Asset Management', `Access Control', `Cryptography', `Physical and environmental', `Operations', `Communications', `System development and acquisition', `Supplier relations', `Information security incident management', `Compliance', `Security Policies' and `Security organisation'. These groups and controls are represented as 408 concepts in the security ontology to protect assets from vulnerabilities and threats. The upper ontology of the seed ontology is shown in Fig. \ref{fig:SecUppOnto} and the ontology is available on GitHub. 

These 408 concepts are extracted from security ontology `to query related terms, namely hypernyms and hyponyms from DBpedia. DBpedia contains over 5 million entities, allows querying of semantic relationships, concepts and properties encoded in the form of RDF triples. Typically, the RDF triples (subject-verb-object) in an ontology contain a `verb' relationship between concepts. Verbs are typically domain-specific and unavailable in general-purpose knowledge graphs like DBPedia. Hypernyms and hyponyms that denote `is-a' relationship between concepts, are easily available in DBPedia and widely used in ontologies, making these relations an ideal choice to demonstrate $OntoEnricher$ approach. The SPARQL queries (query \ref{SPARQLQuer}) to extract hypernyms and hyponyms from DBpedia for concepts in information security ontology are :
%

\begin{Verbatim}

SELECT * WHERE
{<http://dbpedia.org/resource/$concept>
<http://purl.org/linguistics/gold/hypernym> 
?hypernyms}
\end{Verbatim}
\begin{alltt}
SELECT * WHERE {?hypernyms 
<http://purl.org/linguistics/gold/hypernym>
<http://dbpedia.org/resource/\$concept>}
\end{alltt}\label{SPARQLQuer}

The extracted terms with SPARQL queries are converted to triples of the form $(a, b, label)$ where $a$ denotes the ontology concept, $b$ denotes the DBPedia term and $label$ determines the DBPedia relation between $a$ and $b$. This leads to a dataset of 97,425 triples. These triples are then curated by three domain experts and authors to mark unrelated terms as `none'. This includes pairs that are not related to the domain and pairs that are not related to each other, as both these cases are not needed for ontology enrichment. In addition, since DBPedia often categorizes ontological instances under `hyponyms', some pairs are separately labelled as `instances' if $b$ is an instance of $a$ or as a `concept' if $b$ denotes the concept of which $a$ is an instance. The terms classified include names of experts, organizations, products and tools, attacks, vulnerabilities, malware, virus and many others. Finally, since the number of `none' pairs (89,820) is significantly higher than the number of `non-none' (7,605) pairs, `none' pairs are sorted in order of increasing similarity. The first 5\% of `none' pairs are filtered out, this is experimentally determined to yield better results. The Table \ref{tab:DBpediaExtract} shows composition of dataset after extraction, curation and filtration.
\begin{table}[h]
\centering
\begin{tabular}{|l|c|}
\hline
Relationship & Count \\ \hline
Hypernymy & 2,939 \\ \hline
Hyponymy & 794 \\ \hline
Instances & 2,685 \\ \hline
Concepts & 1,187 \\ \hline
None & 4,490 \\ \hline
Total & 12,096 \\ \hline
\end{tabular}
\caption{Composition of the Dataset}
\label{tab:DBpediaExtract}
\end{table}
\subsubsection{Stage 2: Creation of Corpus}
Once the training dataset is created, a training corpus to provide linguistic information for all terms in the dataset is extracted. Wikipedia is used as it is moderated and structured for model training. The DBPedia is a part of the Wikipedia project, and therefore assures unambigous articles of all extracted dataset terms. As a first step, all corresponding Wikipedia articles for terms in the dataset are extracted and added to the corpus. In addition, other articles related to the information security ontology domain are also extracted. This is done by comparing Doc2Vec \cite{lau2016empirical} similarity of each article with the Wikipedia article on `Information Security'\footnote{https://en.wikipedia.org/wiki/Information\_security} and then filtering in articles with a similarity score higher than a certain threshold (0.27 after manual validation). This threshold is determined to optimize classification accuracy after a validation with a sample corpus. The two-step filtering yielded a 2.6 GB size information security training corpus.
\subsection{Stage 3: Training $OntoEnricher$}
Training dataset and corpus are parsed to generate various dependency paths to connect each pair of terms provided in the training dataset. Here, `dependency paths' refers to the multi-set of all paths that connect a pair of terms in the training corpus. These paths are encoded as a sequence of nodes, where each node is a 4-tuple of the form $(word, POS$\_$tag, dep$\_$tag, dir)$. The $POS$\_$tag$ and $dep$\_$tag$ denote POS and dependency tags of the word respectively, while $dir$ denotes the direction of the edge connecting it to the next node in that dependency path. The term pairs along with extracted dependency paths between them are passed to $OntoEnricher$ for training.
\begin{figure}
    \centering
    \includegraphics[width=1.0\linewidth,keepaspectratio]{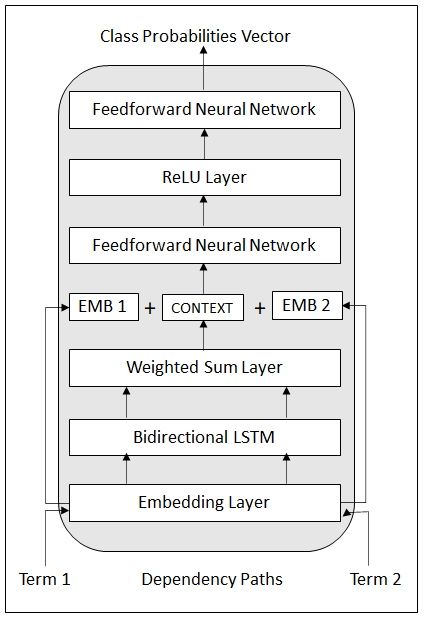}
    \caption{Architecture diagram of OntoEnricher}
    \label{fig:BidirecLSTM}
\end{figure}

The Fig. \ref{fig:BidirecLSTM} shows the architecture diagram of $OntoEnricher$. The first layer in proposed model is the embedding layer. The distributional embeddings for the terms (words) are obtained using a pre-trained state-of-the-art Universal Sentence Encoder (USE) \cite{cer2018universal} model. This model is preferred over other vocabulary-based distributional models such as Word2Vec, Glove and others as it returns distributional embeddings for not just single words, but also compound words, phrases and sentences. In addition, USE is pretrained on Wikipedia along with other corpora, making it suited for this task. Apart from pre-trained word embeddings, embeddings for POS tags, dependency tags and direction tags are obtained from trainable embedding layers. The node embeddings constructed from the concatenation of words, POS, dependency, and direction tag embeddings are arranged in a sequence to obtain path embeddings. A dropout layer is applied after each embedding. The path embeddings for each path connecting the term pair are then input to a bidirectional, two-layer LSTM which trains on a sequence of linguistically and semantically encoded nodes and learns the type of sequences that characterize a particular kind of relation. The bidirectional LSTM allows the network to have both backward and forward information about the path embeddings at every time step, while the two layers enable capturing of complex relations among dependency paths. 

The output of last hidden state of LSTM is taken as the path representation. Since a pair of terms may have multiple paths between them, a weighted sum of these path representations is taken by using path counts as weights, to yield a final context vector. This context vector encodes syntactic and linguistic information, is passed through a dropout layer and then concatenated with distributional embeddings of both terms in order to encode semantic information. The concatenated vector is then passed through two Feedfoward Neural Networks with a Rectifier Linear Unit (ReLU) layer in between, to yield final class probability vector. The class with maximum probability is output as predicted relation between the term pair.
\subsection{Stage 4: Testing $OntoEnricher$}
The procedure to extract concepts and instances from (web page) text, during testing stage is detailed here. To avoid usage of every unstructured (web page) text to enrich an ontology, a lightweight evaluation technique \cite{Sanagavarapu2017ALA} that checks for sufficiency of new security terms is deployed. After passing the sufficiency evaluation as a pre-processing stage, co-reference resolution is applied and then noun chunks are extracted from web page. A cartesian product ($nC_2$) is taken of extracted noun chunks to construct potential term pairs. However, a cartesian product to $OntoEnricher$ is computationally expensive and also leads to error propagation. A two-stage filtering is applied to validate (a) if noun chunks are `sufficiently' related to Information Security and (b) if they are `sufficiently' related to each other. Both these conditions are checked to compare distributional similarity using USE against experimentally determined threshold values. The sufficiently similar term pairs are then input to pre-trained model to classify the relationship. The pairs classified as `None' are discarded and the rest are converted to RDF triples for information security ontology enrichment.
\subsection{Example}
\begin{figure*}
\centering
    \includegraphics[width=1.0\linewidth,keepaspectratio]{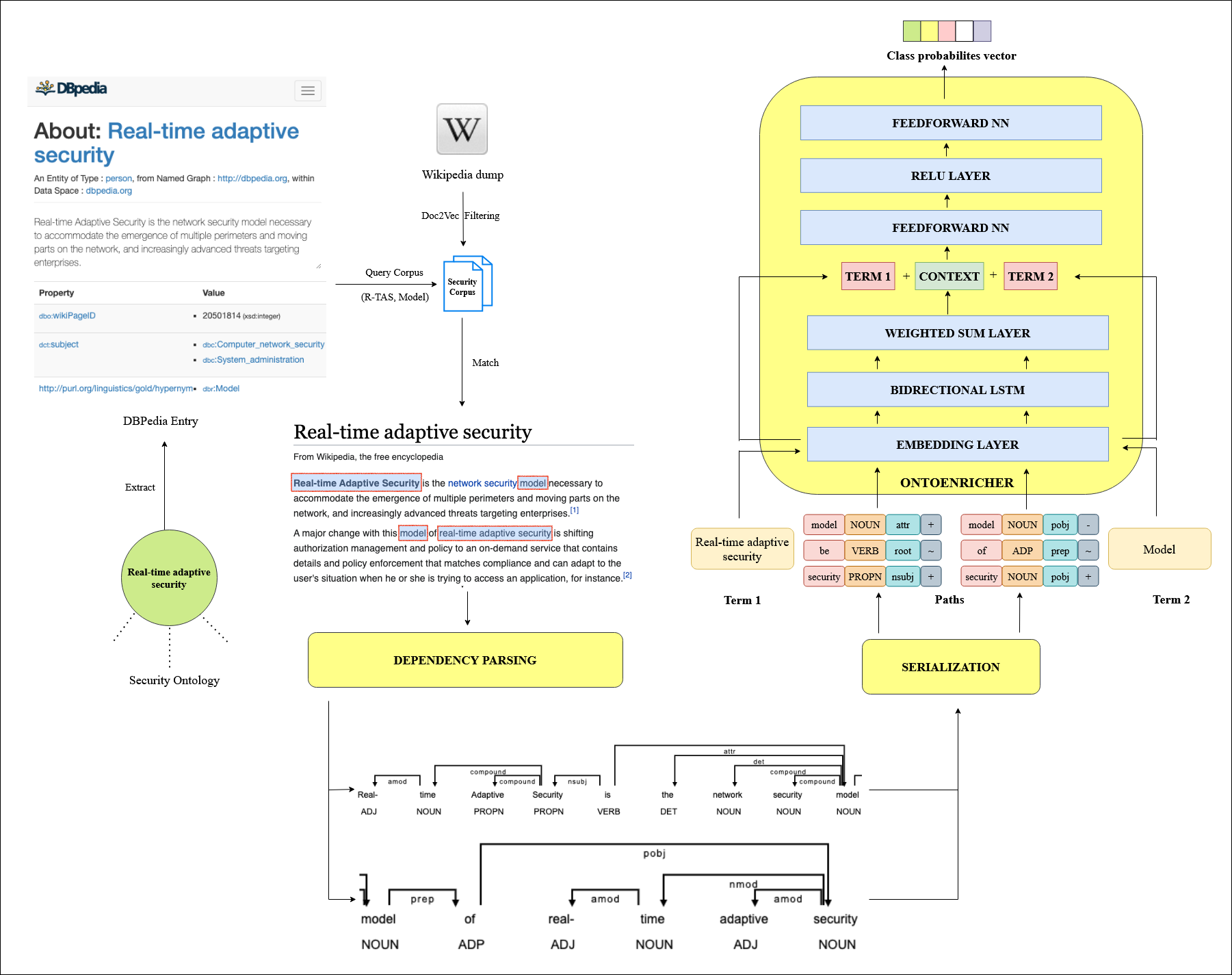}
\caption{Example illustrating Ontology Enrichment approach}
\label{fig:AdaptEnrich}
\end{figure*}

The Fig. \ref{fig:AdaptEnrich} illustrates ontology enrichment approach with an example. `Real-time adaptive security' (R-TAS) is a concept present in information security ontology. The corresponding article in DBPedia, `Real-time adaptive security' has `model' as its hypernymy entry, which is returned using a SPARQL query. The information security corpus extracted from Wikipedia dump using Doc2Vec filter contains multiple paired mentions of these terms, out of which one article contains two mentions. The corpus, the aforementioned sentences, are passed to SpaCy\footnote{https://spacy.io/} dependency parser and all corresponding dependency paths to connect every extracted term pair. These dependency paths which contain encoded linguistic information are passed to a serialization layer that converts the dependency graph into a series of nodes to form the input to $OntoEnricher$. 

The serialization layer reduces the word in every node in the dependency path to its lemma, a root word to enable meaningful training and generalization. Thus, `Real-time adaptive security' is reduced to `security' and `is' is reduced to `be'. It also converts every node to a feature vector. The `Real-time adaptive security' is converted to a feature vector that uses `security' as the word, `PROPN' as POS tag, `nsubj' as dependency tag and `+' denotes the direction of the edge connecting it to the lowest common root node between the term pair. Similarly, the next word `be' is a verb and a root word of `is' does not have any direction `$\sim$'. The last word of this path, `model', has `NOUN' as POS tag, `attr' as dependency tag and `+' as direction of the arrow going away from `model' to `is'. 

The same approach is followed for second dependency path and nodes are sequenced similarly. These two paths are then passed to the embedding layer that calculates (i) USE embedding for word (ii) POS tag embedding (iii) dependency tag embedding and (iv) direction embedding. The last 3 embeddings are trainable while word embeddings are pre-trained using USE. These are concatenated together to yield a node embedding. All paths (node sequences) that connect term pair are passed as input to  Bidirectional two-layer LSTM. In this example, both the paths connecting `Real-Time Adaptive Security' and `model' are input to LSTM, post which the last hidden state is taken as path-wise contextual output. A weighted sum of these paths is then calculated using frequency of occurrence as weights to yield final context vector, this has encoded linguistic information of paths that connect `R-TAS' and `model'. This context vector is concatenated with distributional embeddings of `Real-Time Adapative Security' and `model'. Reducing words to their root form during serialization stage enables to construct a contextualized representation. The characterized paths constitutes a specific relation and the most frequent ones, while distributional word and phrase embeddings enable semantic relevance and specificity at a conceptual level. This concatenated vector denotes semantic and linguistic information that are passed to 2 Feedforward Neural Networks with a ReLU layer in between, yielding a final class probability vector as output. This class probability vector is trained to identify relationship between `model' and `Real-Time Adapative Security' as hypernymy.
\section{Experimental Settings and Results}
The experiment is conducted with two ontologies, namely the ISO 27001-based Information Security and Stanford Pizza ontologies. While the former is focus of this section and use case to build knowledge base, Pizza ontology is used to demonstrate generalizability of the approach. The Table \ref{tab:modelData} shows the composition of security and pizza datasets respectively. While the information security corpus is 2.8 GB in volume, interestingly, the pizza corpus is significantly smaller and only 95 MB. This can be attributed to the fact that the pizza ontology represents a very narrow domain (`pizza' out of food domain) and thus contains few relevant Wiki articles. Information security ontology contains broader, systems-level concepts, information about assets, controls etc. that return a variety of related articles. 

\begin{table}[h]
\centering
\begin{tabular}{|l|l|l|}
\hline
\textbf{Parameters} & \textbf{Security} & \textbf{Pizza} \\ \hline
\# of Concepts & 408 & 143 \\ \hline
Dataset size & 12,096 & 7,119 \\ \hline
Corpus size & 2.8GB & 95MB \\ \hline
\end{tabular}
\caption{Dataset Composition}
\label{tab:modelData}
\end{table}
The $OntoEnricher$ is implemented using deep learning library Pytorch with `0' as random seed number for consistency in results. Also, various other Python libraries such as Pronto\footnote{https://pypi.org/project/pronto/} to extract ontology terms, Wikiextractor\footnote{https://github.com/attardi/wikiextractor} to extract articles from Wikipedia dump, spaCy for dependency graph extraction, and Tensorflow-Hub to load Universal Sentence Encoder are used. The deployed HPC expedites training and testing performance of $OntoEnricher$, this also aids in parallel processing of adding or retrieving concepts, relations and instances from ontology. The performance of $OntoEnricher$ is evaluated on three diverse test datasets :
\begin{enumerate}
     \item DBPedia test dataset: This is created by randomly extracting 10\% of the training dataset extracted from DBpedia. It mostly consists of small-medium length words.
    \item `Knocked-out' test dataset: This is created by knocking out concepts and relations from the seed ontology. This evaluates the ability of $OntoEnricher$ to identify multi-word or phrase-level concepts, as is common in information security ontology, and identification of highly-domain specific, non-English terms as in Pizza ontology.
    \item Instance dataset: This is created by extracting text from security-domain related webpages. The Top 10 vulnerability related web pages from OWASP and product pages on `firewall' are extracted to test the model. The ability to identify concepts and instances from web pages confirms that $OntoEnricher$ can use text from public forums and other unstructured data sources. This evaluation is done without factoring sufficiency requirement \cite{Sanagavarapu2017ALA} of new terms in text to evaluate identification of ontology terms by $OntoEnricher$.
\end{enumerate}

The Table \ref{tab:modelHyp} shows optimized hyperparameters after tuning $OntoEnricher$. Grid search is used to experiment with and arrive at optimal values of various hyperparameters. It includes hidden dimensions (120, 180, 200, 250, 300, 500, 900), input dimension of 2nd NN (60, 90, 120, 180, 300, 500), number of LSTM layers (1,2), activation functions (Softmax, ReLU, LogSoftmax), loss functions (NLL Loss, Cross Entropy loss), and learning and weight decay rates (0.001, 0.01). The experimentation data with various embeddings, epochs, learning rate, activation functions, hidden layers and the related results are available as spreadsheet on GitHub.
\begin{table}[h]
\centering
\begin{tabular}{|l|l|l|}
\hline
\textbf{Hyperparameters} & \textbf{Security} & \textbf{Pizza} \\ \hline
Activation Function & Log Softmax & Log Softmax \\ \hline
Number of Layers & 2 & 2 \\ \hline
Hidden  Dimension of LSTM & 180 & 250 \\ \hline
Input Dimension (2nd NN) & 120 & 90 \\ \hline
Embedding layer Dropout & 0.35 & 0.35 \\ \hline
Hidden layer Dropout & 0.8 & 0.8 \\ \hline
Hidden layer Dropout & 0.8 & 0.8 \\ \hline
Optimizer & AdamW & AdamW \\ \hline
Loss function & NLL Loss & NLL Loss \\ \hline
Epochs & 200 & 200 \\ \hline
Learning Rate & 0.001 & 0.001 \\ \hline
Weight Decay & 0.001 & 0.001 \\ \hline
Weight Initialization & Xavier & Xavier \\ \hline
\end{tabular}
\caption{Hyperparameters of the Model}
\label{tab:modelHyp}

\end{table}

The evaluation results of $OntoEnricher$ on information security and pizza ontologies are shown in Tables \ref{tab:SecEnrich} and \ref{tab:PizzaEnrich} respectively. A competent and comparable scores on information security ontology enrichment with all three datasets are achieved. The test results with 10\% test dataset performed better, while test results on knockout concepts or information security related web pages are not far apart, proving that performance did not dip in extraction of phrases, multi-word concepts and instances which is a key component missing from previous ontology enrichment approaches. As input and output format of existing approaches are different, only a qualitative comparison is performed and shown in Table \ref{Tab:OntoEnrichCompare}. Additionally, in $OntoEnricher$, the number of terms and the size of the corpus used for training and testing are much larger. It is observed that the difference between Precision and Recall value is less, indicates that terms are not skewed towards domain and establishes robustness of the proposed $OntoEnricher$ approach.
\begin{table}[h]
\centering
\begin{tabular}{|l|c|c|c|}
\hline
\textbf{\begin{tabular}[c]{@{}l@{}}Metrics\end{tabular}} & \multicolumn{1}{l|}{\textbf{\begin{tabular}[c]{@{}l@{}}DBPedia\end{tabular}}} & \multicolumn{1}{l|}{\textbf{Knocked out}} & \multicolumn{1}{l|}{\textbf{\begin{tabular}[c]{@{}l@{}}Web pages\end{tabular}}} \\ \hline
\textbf{Terms} & 1197 & 5538 & 153 \\ \hline
\textbf{Accuracy} & 0.81 & 0.77 & 0.83 \\ \hline
\textbf{Precision} & 0.76 & 0.84 & 0.84 \\ \hline
\textbf{Recall} & 0.76 & 0.77 & 0.73 \\ \hline
\textbf{F1-Score} & 0.76 & 0.80 & 0.78 \\ \hline
\end{tabular}
\caption{Security Ontology Enrichment Results}
\label{tab:SecEnrich}
\end{table}
\begin{table*}[h]
\huge
\centering
\resizebox{\textwidth}{!}{%
\begin{tabular}{|l|l|l|l|l|}
\hline
\textbf{Approach} & \textbf{Dataset} & \textbf{Model} & \textbf{Evaluation Metric} & \textbf{Observation} \\ \hline
\begin{tabular}[c]{@{}l@{}}Relation between two named entities \\ identified using NER technique \cite{pingle2019relext}\end{tabular} & \begin{tabular}[c]{@{}l@{}}Open source threat intelligence\end{tabular} & Neural Network and Word2Vec & 96\% as accuracy & \begin{tabular}[c]{@{}l@{}}Single words are only handled and there is\\ no base ontology. Evaluation is performed\\ on pre-trained data\end{tabular} \\ \hline
\begin{tabular}[c]{@{}l@{}}Dual Iterative Pattern Relation \\ Expansion for relation extraction \cite{jones2015towards}\end{tabular} & \begin{tabular}[c]{@{}l@{}}Security related articles from \\ web page\end{tabular} & Semi-supervised models & 82\% as accuracy & \begin{tabular}[c]{@{}l@{}}Not enough volume of training data to \\ validate scalability and generalizability. \\ Ontology and dataset details are not available\end{tabular} \\ \hline
\begin{tabular}[c]{@{}l@{}}Named Entity recognition to \\ identify vulnerabilities and relations \cite{mulwad2011extracting}\end{tabular} & \begin{tabular}[c]{@{}l@{}}NVD, DBPedia and other open \\ source threat intelligence\end{tabular} & Support Vector Machines & 90\% as accuracy & \begin{tabular}[c]{@{}l@{}}Only vulnerability related words are handled \\and there is no base ontology. Evaluation \\is performed on pre-trained data\end{tabular} \\ \hline
Malware text classification \cite{manikandan2018teamdl} & MalwareTextDB & \begin{tabular}[c]{@{}l@{}}Convolutional Neural Network and \\ Conditional Random Field\end{tabular} & 25 - 36\% as accuracy & \begin{tabular}[c]{@{}l@{}}Used Glove for word embeddings that is \\ not fully context sensitive\end{tabular} \\ \hline
\begin{tabular}[c]{@{}l@{}}Named Entities are considered as \\ concepts in security content \cite{gasmi2019cold}\end{tabular} & NVD, Microsoft Bulletins & \begin{tabular}[c]{@{}l@{}}Long Short Term Memory and \\ Conditional Random Field\end{tabular} & 96\% as accuracy & \begin{tabular}[c]{@{}l@{}}The dataset was similar and there are no \\ references to handle multi-word and instances\end{tabular} \\ \hline
Entity extraction from DBPedia \cite{exner2012entity} & Wikipedia and DBPedia & \begin{tabular}[c]{@{}l@{}}Semantic Role Labeler and \\ coreference resolution\end{tabular} & 66.3\% as F1 score & \begin{tabular}[c]{@{}l@{}}No base ontology and references to handle \\ multi-words or instances\end{tabular} \\ \hline
\end{tabular}%
}
\caption{Ontology Enrichment Comparison}
\label{Tab:OntoEnrichCompare}
\end{table*}

Interestingly, the pizza enrichment results shown in Table \ref{tab:PizzaEnrich} are better than security enrichment results, presumably due to domain being narrow as mentioned earlier and concepts are easily identifiable as a consequence.
\begin{table}[h]
\centering
\begin{tabular}{|l|c|c|c|}
\hline
\textbf{\begin{tabular}[c]{@{}l@{}}Metrics\end{tabular}} & \multicolumn{1}{l|}{\textbf{\begin{tabular}[c]{@{}l@{}}DBPedia\end{tabular}}} & \multicolumn{1}{l|}{\textbf{Knocked out}} & \multicolumn{1}{l|}{\textbf{\begin{tabular}[c]{@{}l@{}}Web pages\end{tabular}}} \\ \hline
\textbf{Terms} & 791 & 85 & 99 \\ \hline
\textbf{Accuracy} & 0.99 & 0.79 & 0.88 \\ \hline
\textbf{Precision} & 0.81 & 0.99 & 0.84 \\ \hline
\textbf{Recall} & 0.91 & 0.79 & 0.81 \\ \hline
\textbf{F1-Score} & 0.86 & 0.88 & 0.82 \\ \hline
\end{tabular}
\caption{Pizza Ontology Enrichment Results}
\label{tab:PizzaEnrich}
\end{table}

\begin{figure}[h]
\centering
    \includegraphics[width=1.0\linewidth,keepaspectratio]{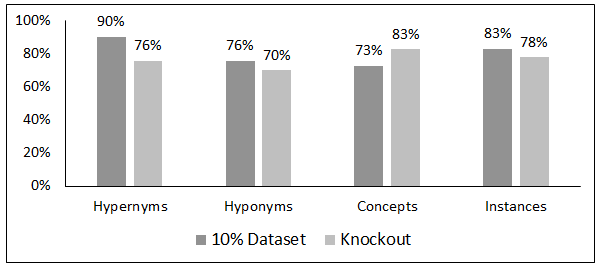}
\caption{Accuracy on Class Identification in Information Security Ontology}
\label{fig:ClassAccuracy}
    \vspace{-3mm}
\end{figure}

Most of the existing ontology evaluation metrics \cite{sabou2005learning} are extensions of Precision and Recall information retrieval metrics. Hence, precision score for $k$ documents (shown in Table \ref{tab:Sec-Precision}) is measured to validate consistency in ontology enrichment with webpages. The scores indicate that the proposed approach can identify concepts for any large number of domain documents. The Fig. \ref{fig:ClassAccuracy} shows the relationship accuracy for each of the classes. It is observable that all relationships are classified equally and hypernymy classification seems to be relatively higher.
\begin{table}[h]
\centering
\begin{tabular}{|l|c|c|c|c|}
\hline
\textbf{Web pages} & \textbf{P@5} & \textbf{P@10} & \textbf{P@15} & \textbf{P@20} \\ \hline
Score              & 0.89         & 0.80          & 0.82          & 0.84          \\ \hline
\end{tabular}
\caption{Precision scores for 20 Random Web Pages in Information Security}
\label{tab:Sec-Precision}
\end{table}
\section{Conclusion and Future Work}
The implemented Information security ontology enrichment approach is comprehensive with an ability to handle new terms, changing domain content that includes concepts, relations and instances. Usage of well accepted ISO 27001 based security ontology, an exhaustive data source such as DBpedia and Wikipedia, Universal Sentence Encoder for distributional embeddings and Bidirectional LSTM for sequential learning makes it extensible to other domains as well. In the implemented enrichment approach, concepts in seed ontology can be a single or multiple words, is an improvement from state-of-the-art. The approach also incorporated instances from unstructured text (web pages) so that organizations or individuals have flexibility to reason information security ontologies for mitigation strategies, vulnerabilities assessment, attack graphs detection and many other use cases. The enriched security ontology can also be used by search engines to display relevant results, top trends in vulnerabilities, threats, attacks and controls. The implemented $OntoEnricher$ is trained on 408 Information Security ontology terms, 97,425 DBpedia terms and 2.8 GB size Wikipedia articles with a HPC cluster. The $OntoEnricher$ is tested with 20 random information security related web pages extracted from internet with an accuracy of 80\% and an F1-score of 78\%. While state-of-the-art results are achieved implementing, the following activities are being explored as future work - 
\begin{itemize}
\setlength{\itemsep}{0pt}
\setlength{\parskip}{0pt}
\setlength{\parsep}{0pt}
    \item Optimize effort required to create DBPedia dataset such as filtering out irrelevant terms.
    \item Test the approach with other security ontologies and extend training corpus beyond Wikipedia. 
    \item Compare results with other knowledge graph and ontology enrichment approaches after curation of input and output format of dataset and corpus.
    \item While there is a need for domain experts to evaluate an enriched ontology, it is effort intensive and brings in other dependencies. A syntactic and semantic evaluation with a easily configurable rules and AI models to reduce effort.
\end{itemize}
\balance
\bibliographystyle{IEEEtran}
\bibliography{references.bib}
\end{document}